\documentclass[sigconf,nonacm]{acmart}
\setcopyright{none}
\usepackage{enumitem}
\usepackage{caption}

\author{Zhen Xu}
\email{zx2393@tc.columbia.edu}
\affiliation{
  \institution{Columbia University}
  \country{USA}
}

\author{Vedant Khatri}
\affiliation{
  \institution{University of California, Irvine}
  \country{USA}
}
\author{Yijun Dai}
\affiliation{
  \institution{Columbia University}
  \country{USA}
}
\author{Xiner Liu}
\affiliation{
  \institution{University of Pennsylvania}
  \country{USA}
}
\author{Siyan Li}
\affiliation{
  \institution{Columbia University}
  \country{USA}
}
\author{Xuanming Zhang}
\affiliation{
  \institution{Columbia University}
  \country{USA}
}
\author{Renzhe Yu}
\email{renzheyu@tc.columbia.edu}
\affiliation{
  \institution{Columbia University}
  \country{USA}
}

\begin{document}
\title{Enhancing LLM-Based Data Annotation with Error Decomposition}

\begin{abstract}
Large language models (LLMs) offer a scalable alternative to human coding for data annotation tasks, enabling the scale-up of research across data-intensive domains such as learning analytics. While LLMs are already achieving near-human accuracy on objective annotation tasks, their performance on subjective annotation tasks, such as those involving psychological constructs, is less consistent and more prone to errors. Standard evaluation practices typically collapse all annotation errors into a single alignment metric, but this simplified approach may obscure different kinds of errors that affect final analytical conclusions in different ways. Here, we propose a diagnostic evaluation paradigm that incorporates a human-in-the-loop step to separate task-inherent ambiguity from model-driven inaccuracies and assess annotation quality in terms of their potential downstream impacts. We refine this paradigm on ordinal annotation tasks, which are common in subjective annotation. The refined paradigm includes: (1) a diagnostic taxonomy that categorizes LLM annotation errors along two dimensions: source (model-specific vs. task-inherent) and type (boundary ambiguity vs. conceptual misidentification); (2) a lightweight human annotation test to estimate task-inherent ambiguity from LLM annotations; and (3) a computational method to decompose observed LLM annotation errors following our taxonomy. We validate this paradigm on four educational annotation tasks, demonstrating both its conceptual validity and practical utility. Theoretically, our work provides empirical evidence for why excessively high alignment is unrealistic in specific annotation tasks and why single alignment metrics inadequately reflect the quality of LLM annotations. In practice, our paradigm can be a low-cost diagnostic tool that assesses the suitability of a given task for LLM annotation and provides actionable insights for further technical optimization.\footnote{Data and code to replicate this paper and apply our method to other annotation tasks can be accessed at: \url{https://github.com/AEQUITAS-Lab/Data-Annotation-LAK-2026}} 

\end{abstract} 


\maketitle

\section{Introduction}
Data annotation, the process of labeling raw data with relevant information~\cite{tan2024large}, is a foundational step in data-driven research. By transforming unstructured data into structured labels, data annotation enables subsequent quantitative analysis, model development, and interpretation. Historically, this process has been labor-intensive, requiring trained human annotators with domain-specific knowledge. Recent advances in generative AI have prompted a significant paradigm shift, as large language models (LLMs) can now perform annotation at scale. Across a wide range of application domains, LLMs have demonstrated near-human annotation performance on tasks such as relation extraction~\cite{wadhwa2023revisiting}, topic classification~\cite{aldeen2023chatgpt}, and text annotation~\cite{gilardi2023chatgpt}. These successes have sparked growing interest in extending LLM annotation to more interpretive tasks in the social sciences. A growing number of researchers have adopted LLMs to accelerate qualitative coding~\cite{ziems2024can,cohn2024towards,grossmann2023ai}, seeking to reduce costs and scale the scope of empirical studies that were previously constrained by limited human resources.

However, using LLMs for social science annotation tasks has turned out to be much harder than expected. For example, studies in learning analytics have used LLMs for complex annotation tasks such as essay scoring, emotion detection, curriculum analytics, and classroom discourse coding~\cite{pack2024large,xu2025course,tanaka2024leveraging,hu2024enhancing} and report highly variable performance. In response, researchers have explored a range of technical advancements, including prompt optimization, model fine-tuning, hybrid human–AI workflows, and multi-agent systems, to improve annotation quality~\cite{ashwin2023using,chew2023llm,siyan2025bringing,simon2025comparing}. However, no single technical optimization consistently outperforms others, and in many cases, more complex methods or advanced models yield only marginal performance gains~\cite{liu2025qualitative,zambrano2023ncoder}. Taken together, these findings invite a fundamental reconsideration: are we confronting technical limitations, or are we encountering annotation tasks that are inherently difficult to automate? 

Motivated by this, we step back from model- and method-level comparisons and instead examine annotation through the lens of task characteristics. Broadly, annotation tasks can be divided into two categories: objective annotation and subjective annotation. Objective annotation tasks, such as tagging objects in images or identifying named entities, rely on concrete, verifiable features and assume the existence of objective ground truths. By contrast, many annotation tasks in the social sciences involve subjective annotations that require annotators to infer latent or theory-driven concepts from contextual cues. These tasks are inherently more interpretive and context-dependent. In traditional social science research, such tasks are commonly framed as content analysis~\cite{krippendorff2018content}, a well-established qualitative methodology. Within this paradigm, complete objectivity and perfect agreement are not realistic expectations when coding psychological constructs, even among trained human annotators~\cite{o2020intercoder}. Theoretically, disagreement in these settings reflects the inherent nature of construct measurement rather than annotator error. From this perspective, the error of LLM annotation in those subjective annotation tasks may stem not only from technical limitations but also from the intrinsic ambiguity of the construct itself.

However, under the current standardized evaluation paradigm, which assesses LLM annotation primarily by its alignment with human-labeled ground truth, metrics such as Cohen’s kappa or accuracy collapse all errors into a single score, offering limited insight into the underlying sources of disagreement or how annotation quality might be effectively improved. To address this limitation, we situate subjective LLM-based annotation within the methodological framework of content analysis, where disagreement is not just treated as an error but as a signal reflecting both task-inherent ambiguity and annotator-specific variability. Building on this perspective, we propose a diagnostic evaluation paradigm that extends conventional LLM annotation workflows with a human-in-the-loop step to estimate task-inherent ambiguity. This mechanism enables evaluation that separates irreducible task ambiguity from model-driven errors and to assess annotation quality in terms of its potential downstream impact. While this diagnostic mechanism is intended to apply broadly to subjective annotation tasks, operationalizing downstream impacts requires task-specific instantiations. As an initial step, we focus on ordinal annotation tasks in this study, where errors naturally vary in severity and downstream consequences, providing a clear and interpretable testbed for validating our proposed mechanism. Specifically, we proceed in three steps:

\begin{enumerate}
    \item Propose an error diagnosis framework for LLM annotation, which categorizes errors by both source (model-specific vs. task inherent) and type (boundary ambiguity vs. conceptual misidentification).
    \item Introduce a lightweight human annotation test to serve as a layer for estimating task-inherent ambiguity within existing LLM annotation workflows.
    \item Develop a statistical method for error decomposition that estimates the distribution of LLM annotation errors by distinguishing their sources and types.
\end{enumerate}

We validate our proposed diagnostic evaluation paradigm on four publicly available educational annotation tasks, demonstrating both its conceptual validity and practical utility. Our contribution to existing research and practice is threefold. First, we provide empirical evidence that explains the widely observed variability in LLM annotation performance on social science tasks and clarifies why very high alignment should not always be expected in inherently ambiguous tasks. Second, we demonstrate the limitations of alignment-based evaluation, showing that for inherently ambiguous tasks, pushing alignment beyond the upper bound imposed by task-inherent ambiguity can risk overfitting noise and subjective biases in human-labeled data. Additionally, single alignment scores could obscure differences in construct capture and downstream impact. Third, we introduce a low-cost diagnostic approach for assessing task suitability for LLM-based annotation and identifying when further technical optimization is likely to be effective. Together, our work advances a paradigm shift in the evaluation of subjective LLM annotation, moving beyond alignment-driven optimization toward a diagnostic approach that prioritizes downstream validity and supports more interpretable and trustworthy use at scale.

\section{Related Work}
\subsection{LLM-Based Data Annotation}
Traditionally, the creation of labeled datasets relies on either high-cost expert annotation or scalable crowdsourcing platforms. The rapid integration of LLMs into annotation workflows represents one of the most significant recent methodological shifts in domains like machine learning, computational social sciences, and other scientific disciplines. Early studies have evaluated LLM performance on a range of annotation tasks and show that, for many relatively objective tasks, such as named entity recognition, span annotation, relation extraction, and natural language inference, LLMs can match or even outperform human crowdworkers\cite{He2023annollm,wang2023gpt,aguda2024large,kasner2025large,tornberg2023chatgpt}.  However, for more subjective tasks, particularly in social science domains, LLMs have also been found to struggle and to exhibit greater performance variance~\cite{siyan2025bringing,kumar2025large,liu2025qualitative,gu2025toward,mcclure2024deductive,zambrano2023ncoder,liu2024assessing,tran2024multi,wu2025unveiling,bewersdorff2023assessing}.  

Some studies further investigate systematic biases and error patterns in LLM-based annotation. Prior work has identified biases that are associated with model architecture and training objectives. For example, LLMs exhibit positional bias~\cite{wu2025emergence}, such as favoring earlier options, a tendency linked to the autoregressive decoding process of transformer architectures. They also display verbosity bias~\cite{saito2023verbosity}, systematically preferring longer and more detailed options, likely introduced by reinforcement learning from human feedback (RLHF), where response length is often implicitly rewarded. In parallel, other studies highlight domain-specific challenges in LLM annotation. General-purpose models have been found to struggle in specialized or high-context domains, particularly when annotation requires understanding domain-specific vocabularies or attending to subtle contextual cues needed for fine-grained distinctions~\cite{dominguez2024lawma}. Additional recurring error patterns include difficulty distinguishing among a growing number of annotation categories~\cite{lu2024mitigating} and systematic inconsistencies in retaining or discarding tokens in classification tasks~\cite{lu2024mitigating,nasution2024chatgpt}.

Recent work shifts the focus from annotation accuracy to how LLM-based annotation is evaluated and its implications for downstream analysis and scientific conclusions. For example, a study ~\cite{baumann2025large} replicates 37 annotation tasks from 21 published social science studies and shows that a small number of prompt paraphrases can manipulate any result to be statistically significant. Other works~\cite{egami2023using,gligoric2025can} propose evaluation and mitigation approaches that focus on downstream validity, aiming to support stable and meaningful empirical conclusions. Together, these efforts indicate that unexamined reliance on LLM-based annotation may distort empirical findings or enable unintended manipulation of results, underscoring the need for more cautious use and evaluation practices that explicitly consider how annotation behavior propagates into downstream analysis.

\subsection{Errors in Subjective Annotation} 
In the social sciences, content analysis is a well-established approach for analyzing and interpreting meaning from text. Developed within the qualitative research tradition, it aims to generate systematic, replicable, and valid inferences from textual data~\cite{krippendorff2018content}. In the following section, we draw on this approach to examine the sources and types of disagreement in human subjective annotation and use this perspective to inform our interpretation of LLM annotation behavior.

\subsubsection{Error Sources}
In qualitative research, complete agreement in coding psychological constructs is widely recognized as unrealistic, because textual meaning depends on annotators’ interpretations~\cite{o2020intercoder}. Prior work ~\cite{krippendorff2008systematic,krippendorff2011agreement} distinguishes two main sources of disagreement: systematic and random. Systematic disagreement occurs when judgments are consistently shaped by factors such as ambiguity in the codebook, a flawed theoretical framework, annotators’ stable biases or idiosyncrasies ~\cite{krippendorff2018content} . Such disagreements are largely unavoidable in practice. Random disagreement, by contrast, reflects unsystematic noise arising from temporary human factors, such as momentary inattention or fatigue.

Later work~\cite {weber2024varierr} similarly distinguishes between true annotation errors and human label variation. Annotation errors refer to labels assigned for invalid reasons, such as misunderstanding the instructions or making accidental mistakes. In contrast, human label variation arises when different annotators assign different but still valid labels to the same item due to unavoidable factors. Two main sources of such variation have been identified. The first is linguistic ambiguity, an inherent property of text data in which linguistic entities admit multiple valid interpretations ~\cite{de2012did,manning2011part,plank2014linguistically,beck2020representation}. And the second source is annotator subjectivity, which is particularly pronounced in highly subjective tasks like hate speech detection and sentiment analysis ~\cite{talat2016you,sap2021annotators,poria2020beneath}. 

\subsubsection{Error Types}
In annotation practice, labels serve as inputs to downstream statistical analysis or model development. As a result, the impact of annotation errors depends not only on whether an error occurs, but also on its severity and downstream consequences. This distinction is particularly important for ordinal annotation tasks, where categories are ordered, and misclassifications vary in how strongly they violate the underlying construct~\cite{krippendorff2004measuring} ~\cite{cohen1968weighted}. 

In this context, prior work distinguishes between two conceptually and statistically different types of ordinal annotation errors: adjacent errors and distant errors: adjacent errors and distant errors ~\cite{stevens1946theory,cohen1968weighted,krippendorff2018content,mckelvey1975statistical} ~\cite{agresti2025categorical,berchez2025dlordinal}. Adjacent errors occur when an annotation falls into a neighboring category of the true label, typically reflecting minor boundary disagreements between closely related levels of a construct. This type of error is generally considered less harmful and poses lower risks for downstream analyses~\cite{he2022active}. Distant errors, in contrast, involve misclassifications across non-adjacent levels of the ordinal scale. These errors represent more substantial violations of the construct’s intended ordering and often indicate deeper problems, such as major misinterpretations of the data or systematic misapplication of the annotation framework~\cite{amigo2020effectiveness,triguero2016labelling,cardoso2011measuring}. As a result, distant errors can severely compromise the validity of annotated datasets, distort empirical findings, and introduce misleading signals into model training~\cite{klie2023annotation}. Reflecting this asymmetry in risk, studies of ordinal data commonly rely on evaluation metrics such as weighted kappa or Mean Squared Error to penalize distant errors more heavily than adjacent ones~\cite{vanbelle2024comprehensive,baccianella2009evaluation}.

\section{An Error Decomposition-Based Evaluation Paradigm}
This section is organized as follows. We first outline a conceptual framework for decomposing LLM annotation errors by type and source. Next, we estimate task-inherent ambiguity by introducing a lightweight human-annotation test. Finally, we present a computational method to estimate the distribution of error types and sources in LLM annotations.

\begin{figure*}[h]
  \centering
  \includegraphics[width=0.8\linewidth]{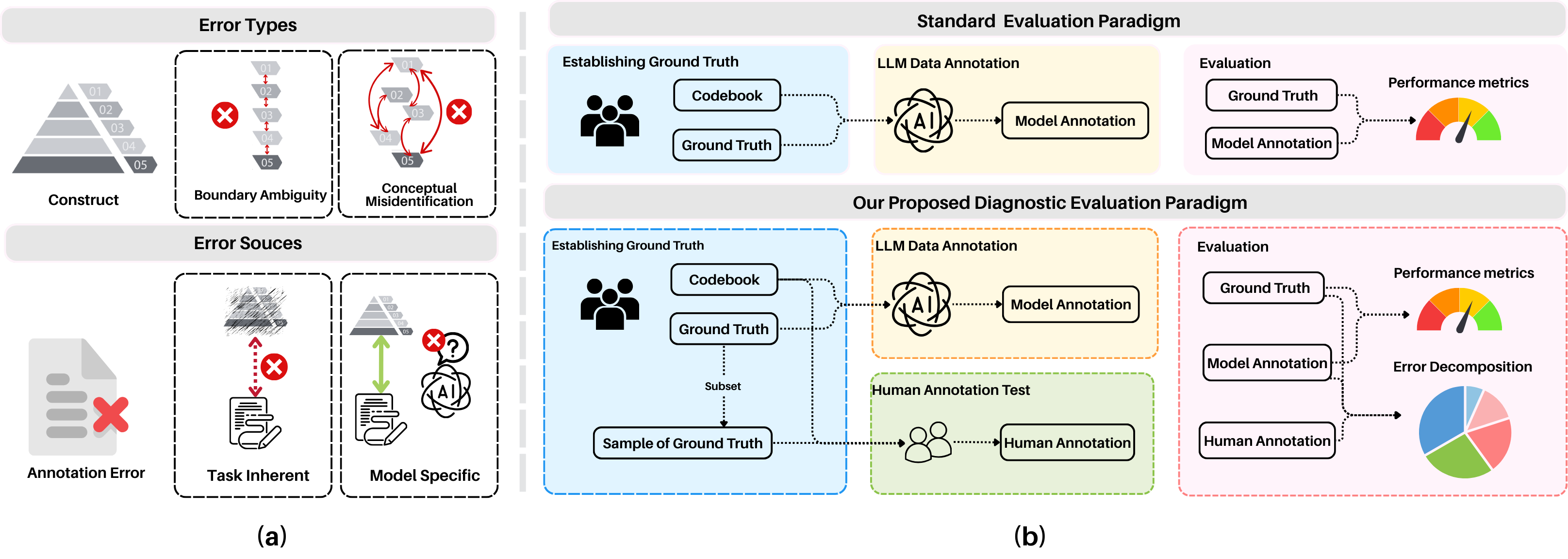}
  \captionsetup{width=0.95\linewidth}
  \caption{\textbf{LLM Subjective Annotation}:  (a) Error Decomposition Framework and (b) Comparison between the standard and our proposed diagnostic evaluation paradigms}
  \label{fig:workflow}
  \Description{}
\end{figure*}
\subsection{Conceptualization of Annotation Errors}

\subsubsection{Error Sources}
Adapting from the content analysis approach, we distinguish annotation error into two sources: \textbf{systematic error (task-inherent)} and \textbf{random error (model-specific and human-specific)}, as illustrated in Figure 1(a). Systematic errors arise from task-inherent ambiguity and are largely unavoidable, while random errors, by contrast, are attributed to annotators and can be reduced. In human annotation, random errors may stem from factors such as memory limitations, decision fatigue, lapses in attention, cognitive load, or insufficient domain expertise. In LLM-based annotation, we conceptualize random errors as arising from model-specific factors, such as inadequate tuning or technical instability, which may be mitigated through improved training or decoding configurations.

\subsubsection{Error Types}
To capture differences in the severity of deviation from the intended construct in ordinal annotation tasks, we distinguish two types of error, as shown in Figure 1(a). \textbf{Boundary Ambiguity} refers to misclassifications that occur near the boundaries of adjacent ordinal categories. These errors typically arise from subtle interpretive differences between neighboring classes and represent minor deviations that largely preserve the intended meaning of the annotation. \textbf{Conceptual Misidentification}, by contrast, occurs when misclassifications cross substantive conceptual boundaries and span multiple levels of the coding levels. Such errors often reflect deeper misunderstandings of the underlying construct and pose a substantially higher risk in downstream analyses or model training.

\subsection{Estimating Task-Inherent Ambiguities with Human Annotations Test}
Building on our framing of error sources, we assume that errors arising from task-inherent ambiguity are systematic and therefore likely to occur consistently in both human and LLM annotation. Therefore, we propose using systematic errors made by human annotators as a proxy to estimate the task-driven error of LLM annotation. To operationalize this, we incorporate an additional step into the current standard annotation evaluation workflow. As illustrated in Fig.~\ref{fig:workflow} (b), after establishing the ground truth, we assign the same task instructions used for the LLM to a separate pair of human annotators who did not participate in ground truth construction. However, we acknowledge that this process may theoretically also introduce human random error, which could confound the estimates. To minimize this effect, we incorporated several safeguards in the annotation process. First, annotators were selected based on their relevant domain expertise to ensure a clear understanding of the construct, thereby reducing random errors from insufficient domain knowledge. Second, the task is conducted on a relatively small and manageable sample without time constraints, minimizing the effects of fatigue, memory decay, time pressure, and attentional lapses.

\subsection{Statistical Method for Decomposing LLM Annotation Errors}
To operationalize two error types in ordinal subjective annotation, we define boundary ambiguity error as cases in which the annotated label falls into the neighboring class of the ground truth (ordinal distance = 1) ,and conceptual misidentification error refers to deviations of two or more levels (ordinal distance $\geq$ 2). In this section, we describe our computational approach for estimating these error types and error sources in LLM-based annotation.

\subsubsection{Error Types Formalization}
For each annotation item, let $Y_G$ denote the ground-truth label from $G$, $\hat{Y}_M$ denote the model prediction, and $\hat{Y}_H$ denote the human annotation. We define the ordinal distances from the ground truth as:
\[
\delta_M = |\hat{Y}_M - Y_G|, \quad \delta_H = |\hat{Y}_H - Y_G|,
\]
where $G$ (Ground Truth), $M$ (Model Predictions), and $H$ (Human Annotation Test) are the three annotation sets described above.  Errors are then categorized according to their distance from the ground truth. For model predictions ($M$), we define:
\begin{align*}
\mathbb{P}_{\mathrm{correct}} &= \mathbb{P}(\delta_M = 0), \\
\mathbb{P}_{\mathrm{model\_boundary}} &= \mathbb{P}(\delta_M = 1), \\
\mathbb{P}_{\mathrm{model\_concept}} &= \mathbb{P}(\delta_M \ge 2).
\end{align*}
Similarly, for human annotations ($H$), we define:
\[
\mathbb{P}_{\mathrm{human\_boundary}} = \mathbb{P}(\delta_H = 1), \quad
\mathbb{P}_{\mathrm{human\_concept}} = \mathbb{P}(\delta_H \geq 2).
\]

\subsubsection{Error Sources Formalization}
Then, we propose a method that leverages the ground truth and human annotation test results to estimate the proportions of task-driven and model-specific errors within the two error types. Formally, the prediction outcome of an LLM can be decomposed into five mutually exclusive categories, and their probabilities sum to one:

\[
\mathbb{P}_{\mathrm{correct}} 
+ \mathbb{P}^{T}_{\mathrm{boundary}} 
+ \mathbb{P}^{M}_{\mathrm{boundary}} 
+ \mathbb{P}^{T}_{\mathrm{concept}} 
+ \mathbb{P}^{M}_{\mathrm{concept}} = 1,
\]
where: $\mathbb{P}^{T}_{\mathrm{boundary}}$: task-driven boundary error;  $\mathbb{P}^{M}_{\mathrm{boundary}}$: model-specific boundary error; $\mathbb{P}^{T}_{\mathrm{concept}}$: task-driven concept error;  $\mathbb{P}^{M}_{\mathrm{concept}}$: model-specific concept error.

\paragraph{Step 1: Estimating Task-Driven Error }
To estimate the proportion of model errors that come from task ambiguity, we use a \textit{probability-sharing} method that compares error patterns between humans and the model. The idea is that when both humans and the model make the same type of error, a portion of the model’s error can be attributed to the difficulty of the task itself rather than to model-specific limitations. Using this assumption, we calculate the task-driven components of boundary and concept errors as follows:

\[
\mathbb{P}^{T}_{\mathrm{boundary}} 
= \frac{\mathbb{P}_{\mathrm{human\_boundary}}}
       {\mathbb{P}_{\mathrm{human\_boundary}} + \mathbb{P}_{\mathrm{model\_boundary}}}
  \cdot \mathbb{P}_{\mathrm{model\_boundary}},
\]
\[
\mathbb{P}^{T}_{\mathrm{concept}} 
= \frac{\mathbb{P}_{\mathrm{human\_concept}}}
       {\mathbb{P}_{\mathrm{human\_concept}} + \mathbb{P}_{\mathrm{model\_concept}}}
  \cdot \mathbb{P}_{\mathrm{model\_concept}}.
\]

\paragraph{Step 2: Estimating Model-Specific Error.}
After estimating the task-driven error, we define the model-specific error as the remaining portion of the model’s total error that cannot be attributed to human disagreement. This residual represents errors that are more likely due to limitations in the model’s understanding, reasoning, or representation of the task. We calculate the model-specific components of boundary and concept errors as follows:
\begin{align*}
\mathbb{P}^{M}_{\mathrm{model\_boundary}} 
= \mathbb{P}_{\mathrm{model\_boundary}} - \mathbb{P}^{T}_{\mathrm{boundary}},
, \\
\mathbb{P}^{M}_{\mathrm{concept}} 
= \mathbb{P}_{\mathrm{model\_concept}} - \mathbb{P}^{T}_{\mathrm{concept}}.
\end{align*}

\section{Experiments}
We validate our diagnostic evaluation paradigm on four educational annotation tasks. Specifically, we test the following hypotheses: 
\begin{itemize}
    \item (H1) The two error patterns observed in human annotations align with annotators’ perceptions of the  task ambiguities;
    \item (H2) Advanced models mitigate model-driven errors more effectively than task-driven ones;
    \item (H3) Prompt optimization mitigates model-driven errors more effectively than task-driven errors.
\end{itemize}
In addition, we conduct an exploratory analysis to assess whether our error decomposition method explains variation in LLM performance across tasks and accounts for differences in the effectiveness of prompt engineering.

\subsection{Annotation Tasks and Ground Truth Datasets}
We use four single-labeled, ordinal datasets in education, each meeting the following criteria: (1) annotated by at least two human annotators, with discrepancies resolved; (2) reported inter-rater reliability (IRR); and (3) the codebook and annotation guidelines are open to the public. From each dataset, we stratify a sample of 150 annotated data points to serve as the ground truth, with an equal number of annotations per class to avoid class imbalance in subsequent performance evaluation. A summary of the datasets is provided in Table~\ref{tab: datasets}.

\begin{table*}[h]
    \centering
    \begin{tabular}{lc}
    \toprule
    Dataset & Task Description\\
    \midrule
    Bloom~\cite{li2022automatic}& Classify 6 learning objectives according to Bloom's Taxonomy. (6 levels)\\
    MathDial~\cite{macina2023mathdial}& Classify teaching moves within tutoring dialogues. (4 levels)\\
    Uptake~\cite{demszky2021measuring}& Rate the extent to which a teacher’s response builds on a student’s sentence.(3 levels)\\
    GUG~\cite{heilman2014predicting}& Rate the grammatical correctness of sentences written by second-language learners. (4 levels)\\
    \bottomrule
    \end{tabular}
    \caption{Four Data Annotation Tasks}
    \label{tab: datasets}
\end{table*}

\subsection{Technical Setup for LLM Annotation}
We adopt five prompting strategies, along with a prompt optimization method that relies on fewer than 300 human-labeled examples, reflecting common practice in low-resource annotation settings. To study how model version affects performance, we test both GPT-3.5 and GPT-5, which represent different generations of models from the same vendor.

The five prompting strategies we adopt are: \textit{zero-shot}, where the model receives only the annotation codebook; \textit{few-shot}, which adds a small number of labeled examples; and \textit{chain-of-thought prompting}, which builds on the zero-shot setup by adding a reasoning cue (“Please think step by step before selecting the label”). We also include \textit{self-consistency}, where the model generates five outputs for each instance, and the most frequent prediction is used (an approach also referred to as ‘majority vote’ in previous studies), and \textit{active prompting}, where challenging examples identified through disagreement and error rates in a 20-sample pool are selected to guide the model on ambiguous cases. Detailed prompts and parameter settings are provided in the Appendix. We also conducted prompt optimization to simulate scenarios where more human-annotated ground truth is available for prompt calibration and refinement. This process was implemented as a classification task under the DSPy framework ~\cite{khattab2023dspy}. Using DSPy v3, we performed optimization with varying amounts of annotated data (100, 200, and 300 samples). 

\subsection{Estimating Task-Inherent Ambiguity with Human Annotation Test}
To probe task-inherent annotation error (Fig.~\ref{fig:workflow}(b)), we recruit two education graduate students with prior annotation and qualitative research experience. Before the annotation test, we held a tutorial session to introduce the annotators to the four tasks and annotation requirements. For each annotation task, the annotators received 40 stratified samples from the ground truth dataset used in our LLM performance evaluation, with balanced class distributions to avoid skew in error type estimation. To maintain comparability between human and model interpretations, the annotators worked independently using the same instruction sets and codebooks provided to the LLMs. To minimize random human error resulting from fatigue, memory load, or loss of focus, the annotation process was conducted in two rounds. Each annotator completed one set of 20 samples at a time, with no time limit, and was encouraged to take breaks between tasks. Both annotators completed each round of the four annotation tasks (20 samples per task) within two days. Inter-rater reliability, measured by Krippendorff’s alpha, is Bloom: 0.872, MathDial: 0.71, Uptake: 0.63, and GUG: 0.60 (Inter-rater reliability originally reported in the corresponding datasets are: Bloom: Cohen’s $k$  = 0.80; MathDial: Cohen’s $k$=0.67; Uptake: Spearman $\rho$= 0.474, and GUG: Cohen’s $\kappa$=0.574. See Appendix for more details.)

After completing the annotation task, annotators individually complete a survey assessing their perceived likelihood of the two types of error for each annotation task. For each error type, we included two direct measurement items: (1) Boundary ambiguity: \textit{The boundary between adjacent categories (e.g., scores 2 and 3) felt clear and distinct} (4 = Most distinct, 1 = Most blurred); (2) Conceptual misidentification: \textit{It was difficult to distinguish between categories that are not adjacent on the scale} (4 = Most difficult, 1 = Easiest). In addition to these direct items, five supplementary items were included for each error type. For each survey item, annotators are asked to rank the four annotation tasks from easiest to most difficult to annotate.

\section{Results}
\subsection{Validation of Error Decomposition Framework}
This session presents results from three hypothetical questions and an exploratory analysis to examine the conceptual validity and practical utility of our error-decomposing method.
\paragraph{\textbf{Do boundary and concept error patterns in human annotations align with human perceptions of boundary clarity and conceptual ambiguity?}}

\begin{figure*}[h]
  \centering
\includegraphics[width=0.32\linewidth]{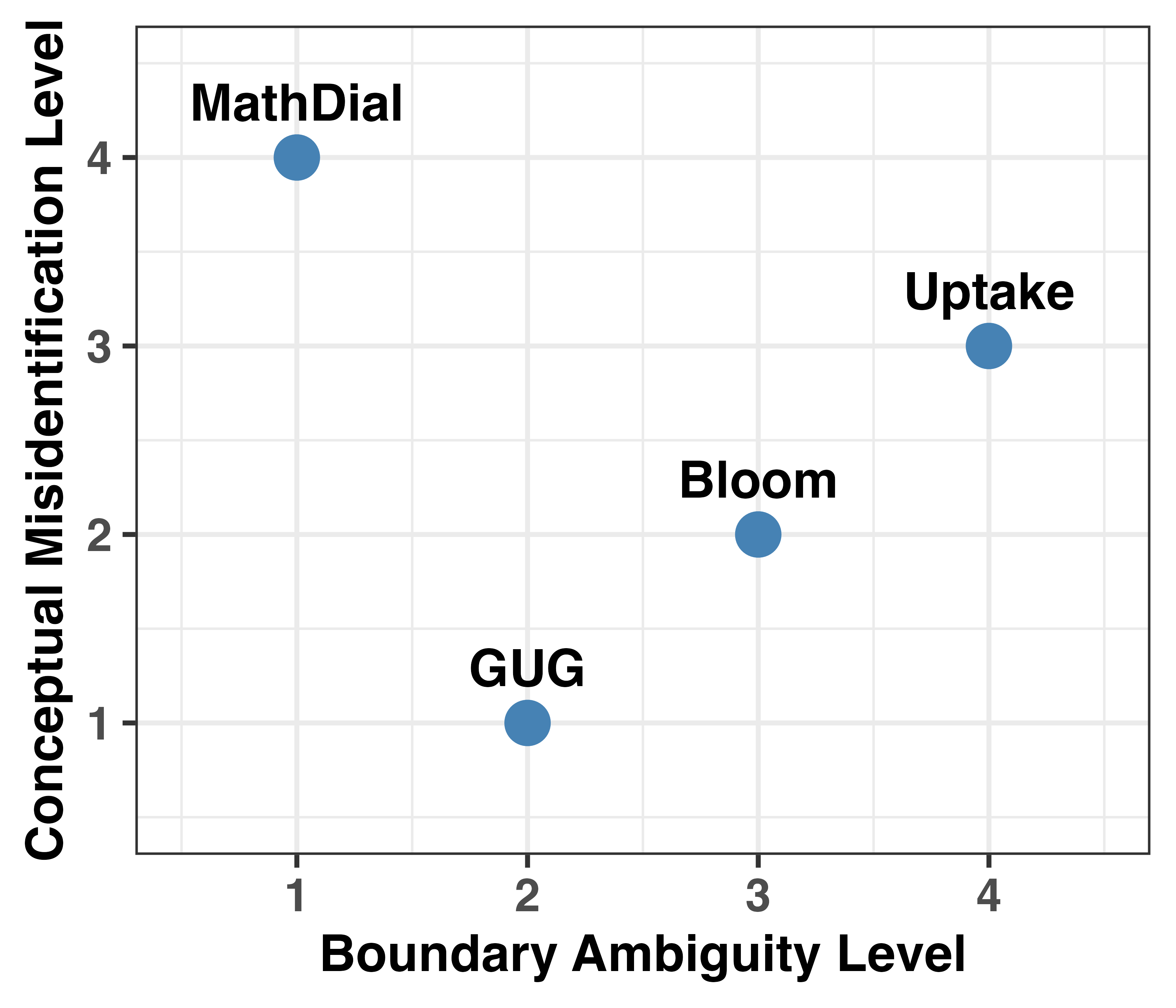}
  \includegraphics[width=0.6\linewidth]{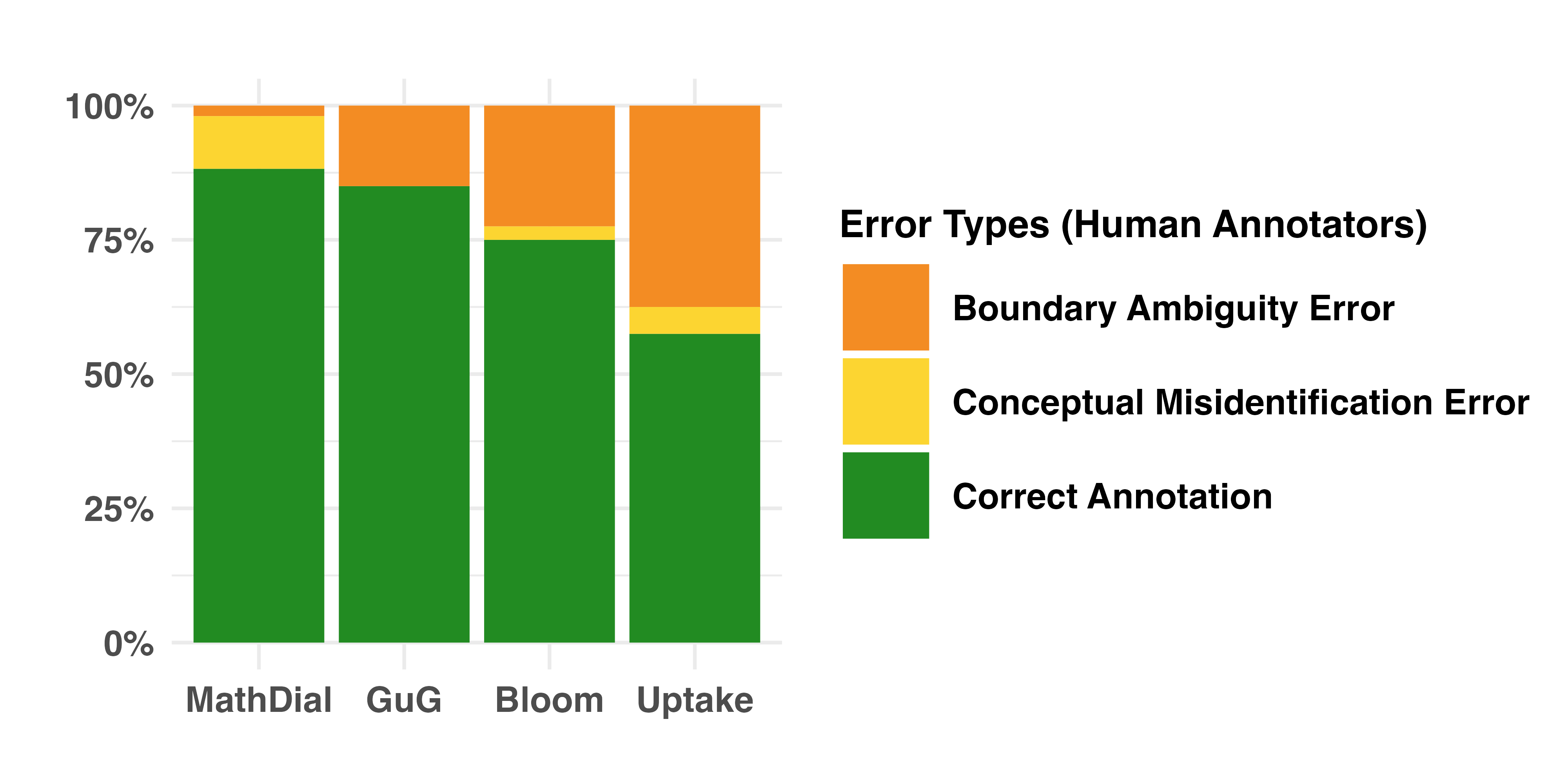}%
  \captionsetup{width=.95\linewidth}
  \caption{\textbf{Human Error Types.} 
  (a) Annotators’ subjective perceptions of boundary vs. construct clarity: Two annotators were asked to rank the four tasks for their boundary and construct clarity level after annotation. Smaller numbers represent easier judgments in both dimensions. 
  (b) Distribution of error types observed in human annotations. }
  \label{fig:human_error}
\end{figure*}

Fig~\ref{fig:human_error}(a) shows the comparison between the annotators’ error patterns observed in their annotations and their subjective perceptions from the post-annotation survey, and the rankings were fully consistent across our two annotators. Figure~~\ref{fig:perform}(b) presents the proportions of boundary errors, concept errors, and correct annotations across the four datasets, relative to the ground truth. We test three sample sizes for the annotation task (20, 30, and 40 items) and found consistent distribution patterns across all settings. For clarity, we report results for the 20-sample settings here and in the following sections; results for the other two conditions are provided in the Appendix.

We observed a clear alignment between subjective perceptions and actual error patterns observed in human annotation: tasks that annotators judged as having a lower boundary or construct clarity also produced higher proportions of the corresponding error types. This alignment provides empirical support for the validity of our two-error typology framework, as annotators’ perceived ambiguity is reflected in the observed distribution of boundary and concept errors.

\begin{figure*}[h]
  \centering
  \includegraphics[width=0.8\linewidth]{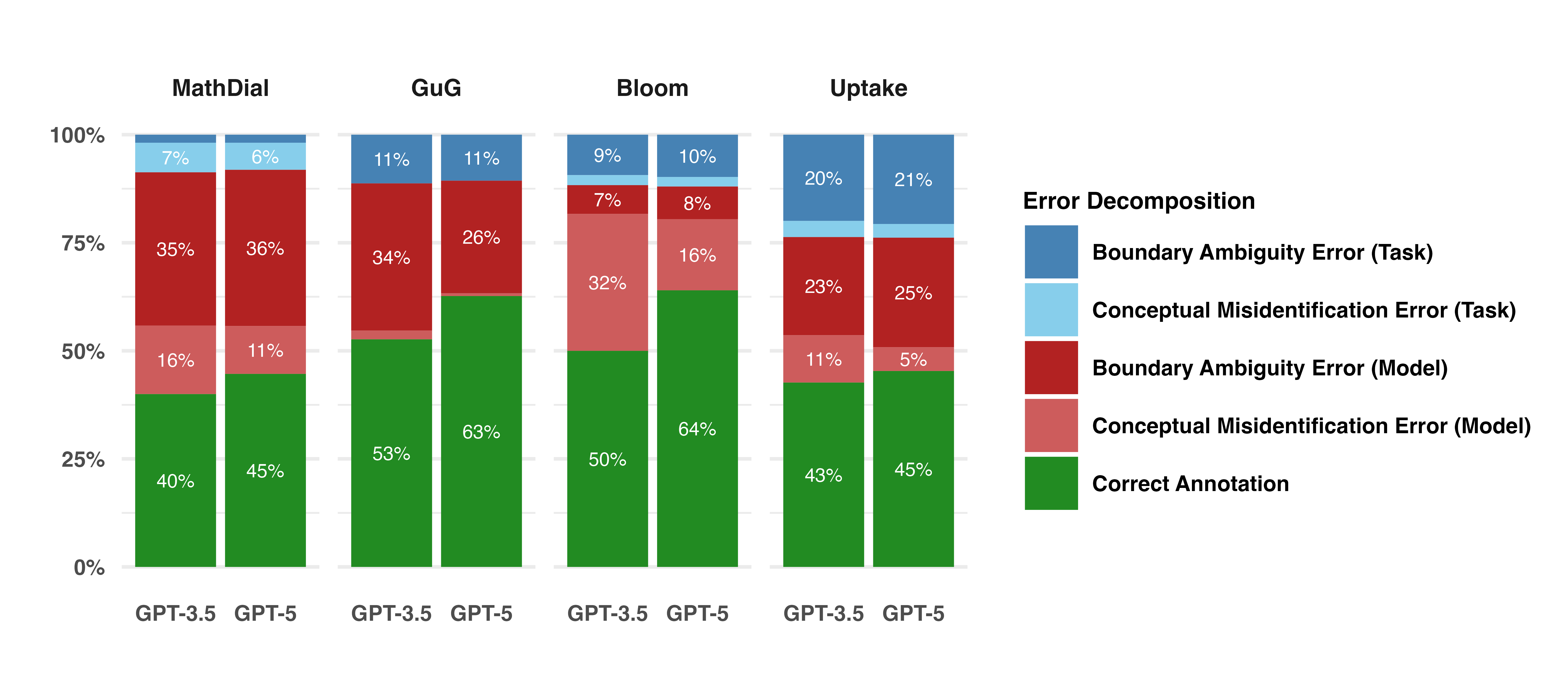}
  \captionsetup{width=.95\linewidth}
  \caption{\textbf{Comparison of Error Decomposition Between Two Models.} LLM (zero-shot) annotation results are shown by error type (boundary vs. conceptual) and source (task-inherent vs. model-specific), with correct annotations in green.}
  \label{fig:model-error}
  \Description{}
\end{figure*}

\paragraph{\textbf{Do advanced models better mitigate model-driven errors than task-driven ones?}}
After we validate the conceptual validity of the two error types, we test whether the proposed error decomposition method can capture distinct sources of error in LLM-generated annotations. We next compare model-level differences under the baseline zero-shot condition, while controlling for prompt design and external guidance to isolate model-driven effects. Figure~\ref{fig:model-error} shows the error decomposition outcomes for GPT-3.5 and GPT-5 across the four annotation tasks in our study.

Across all tasks, we observe that GPT-5 consistently outperformed GPT-3.5 in terms of overall annotation accuracy, and the decomposition shows that this improvement was primarily driven by a reduction in model-specific errors, as illustrated by the red bars in the figure. In contrast, task-driven errors (blue bars) remained mostly stable, with less than a 1\% difference between the two model versions for each of the four annotation tasks we tested. These results support our hypothesis: advanced models reduce model-specific errors and improve alignment with ground truth annotations, but have limited effectiveness in addressing errors rooted in task-inherent ambiguity.

\begin{figure*}[h]
  \centering
  \includegraphics[width=0.8\linewidth]{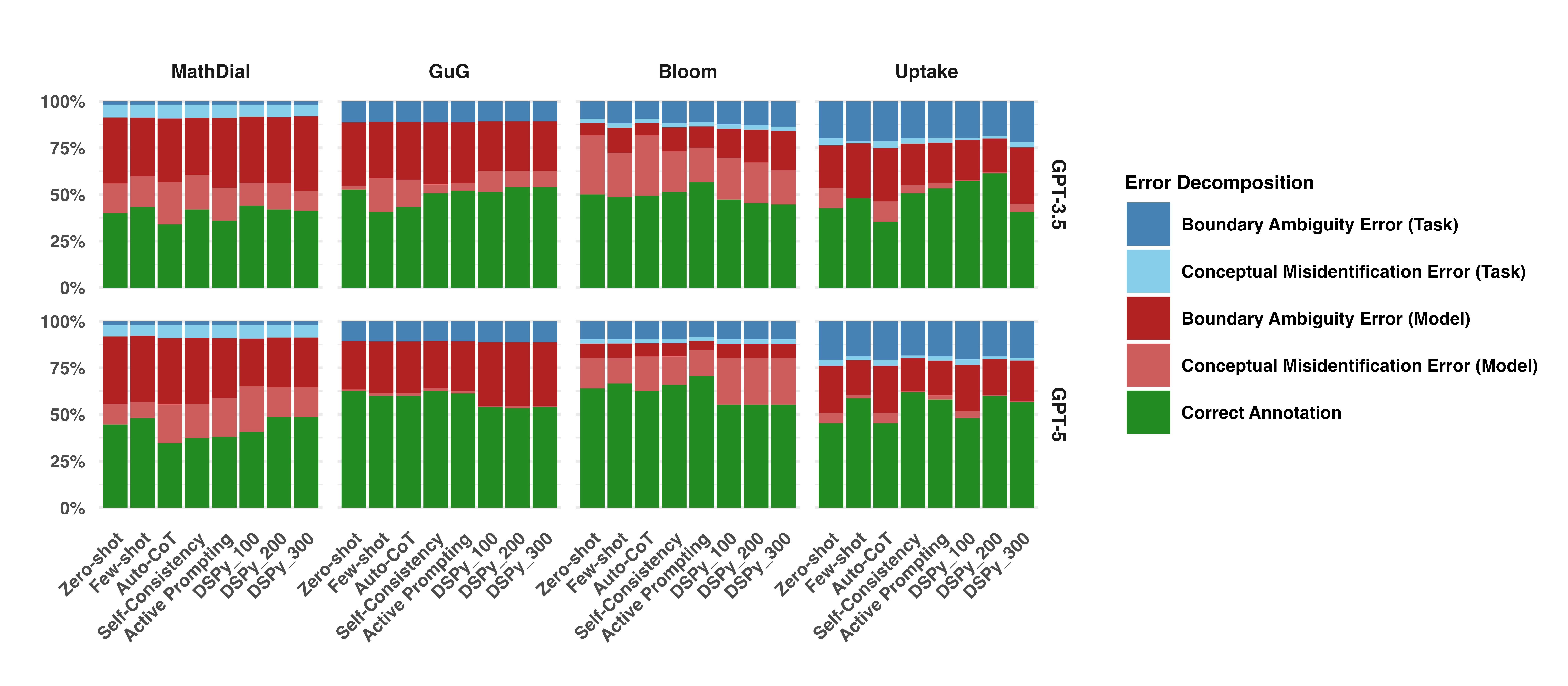}
  \captionsetup{width=0.95\linewidth}
  \caption{\textbf{Error Decomposition Across Prompting Strategies.}LLM annotation results are shown by error type (boundary vs. conceptual) and source (task-inherent vs. model-specific) across different prompting and decoding strategies, with correct annotations shown in green.}
  \label{fig:prompt-error}
  \Description{}
\end{figure*}

\paragraph{\textbf{Does prompting optimization better mitigate model-driven errors than task-driven ones?}}
We next extend the model-level comparison to different prompting strategies. Figure~\ref{fig:prompt-error} shows the error decomposition results across prompting strategies and prompt optimization methods, applied to both GPT-3.5 and GPT-5 for all four annotation tasks.

We observe a consistent pattern of stability in task-driven errors across all prompting strategies. Regardless of the prompting strategies used, or whether optimized prompts incorporated additional human-labeled examples, the proportion of task-driven errors (blue bars in the figure) remained largely unchanged within each task. In contrast, several strategies reduced model-specific errors, showing that prompting can improve alignment with ground truth, but cannot resolve the ambiguity embedded in the task formulation. These results support our hypothesis that targeted prompting strategies reduce model-specific errors but have little effect on errors rooted in task-inherent ambiguity. They also validate our proposed error decomposition method.

Beyond hypothesis testing, we also identify a ceiling effect in the performance of prompting strategies. Even the best-performing methods did not reduce task-driven errors beyond the level observed in the baseline (zero-shot) condition. This result suggests that error decomposition observed under zero-shot conditions can serve as a predictive upper bound on the performance gains achievable through prompting alone. In other words, the proportion of task-driven error in the zero-shot setting likely reflects the irreducible complexity of the task and provides a basis for setting realistic expectations about the limits of prompt-based improvements.

\subsection{Utility of Error Decomposition for Explaining Performance Variation}
To evaluate the usefulness of our proposed error decomposition framework, we conducted an exploratory analysis to test whether task-inherent errors, specifically boundary ambiguity and conceptual misidentification, help explain why some annotation tasks achieve higher baseline performance or benefit more from prompt engineering.

\begin{figure*}[h]
  \centering
  \includegraphics[width=0.8\linewidth]{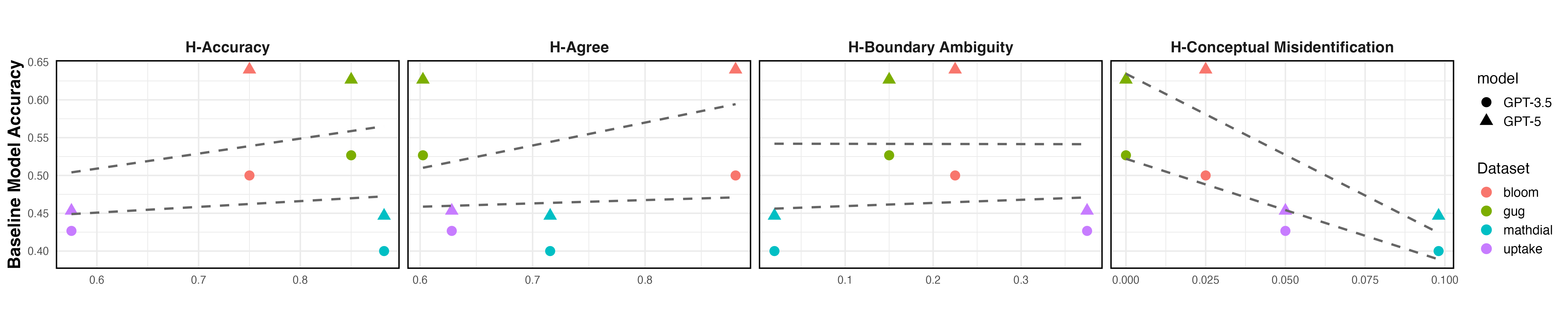}
  \captionsetup{width=.95\linewidth}
  \caption{\textbf{Baseline (Zero-Shot) Model Accuracy vs. Task Inherent Errors.}
 \textit{Baseline Model Accuracy} : the agreement between LLM-generated annotations (zero-shot) and the original dataset labels; \textit{H-Accuracy} : the average agreement between our annotators’ labels and the original dataset labels; \textit{H-Agree}: Krippendorff’s $\alpha$ between our two annotators; \textit{H-Boundary Ambiguity}: boundary ambiguity annotation errors observed between the two annotators;  \textit{H-Conceptual Misidentification}: conceptual misidentification annotation errors observed between the two annotators.}
  \label{fig:baseline}
  \Description{}
\end{figure*}

\paragraph{\textbf{Baseline Model Performance and Task Inherent Errors}}

Figure~\ref{fig:baseline} illustrates how our error decomposition framework explains baseline performance differences across the four educational annotation tasks. The first two panels show overall human annotation accuracy and inter-rater agreement level (Krippendorff's alpha), along with their correlation with LLM performance in the zero-shot setting. Although these metrics are commonly used as proxies for task clarity and complexity, the results reveal no consistent or predictive relationship with model performance. For example, the GUG dataset exhibits moderate inter-rater agreement yet relatively poor LLM performance, whereas Bloom shows both high agreement and strong baseline performance. Such inconsistencies suggest that aggregate measures of agreement or accuracy may be insufficient to capture the task-specific characteristics that shape LLM annotation outcomes. We then apply our framework to further decompose human disagreement. As shown in the fourth panel, tasks with a higher proportion of conceptual misidentification-oriented disagreement by humans are associated with lower baseline model performance. This could suggest that even with the same human agreement level, a high conceptual misidentification could reflect more profound construct-level ambiguity, which may pose a more fundamental challenge for LLMs in zero-shot settings.

\begin{figure*}[h]
  \centering
  \includegraphics[width=0.8\linewidth]{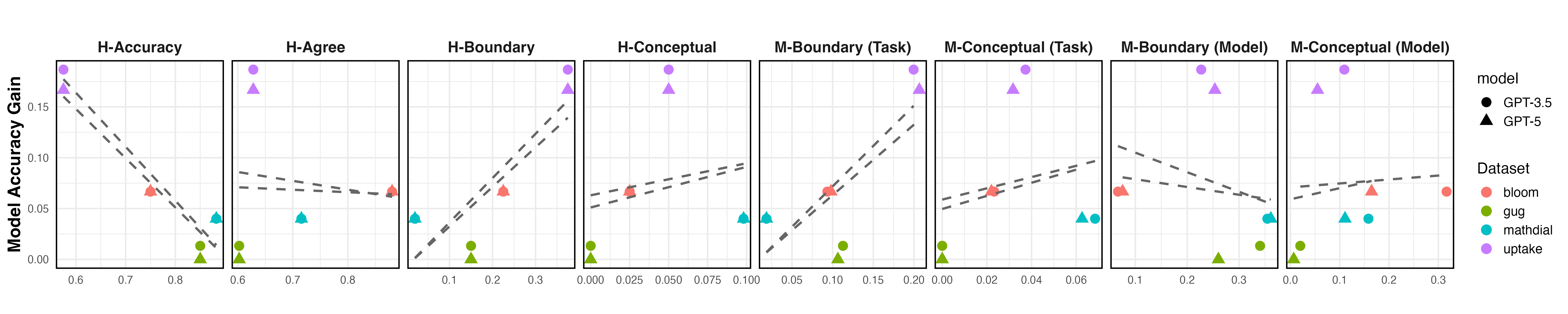}
  \captionsetup{width=.95\linewidth}
  \caption{\textbf{Prompting Strategies Effectiveness vs. Task Inherent Errors}
  \textit{Baseline Model Accuracy} : the agreement between LLM-generated annotations (zero-shot) and the original dataset labels; \textit{H-Accuracy} : the average agreement between our annotators’ labels and the original dataset labels; \textit{H-Agree}: Krippendorff’s $\alpha$ between our two annotators; \textit{H-Boundary Ambiguity}: boundary ambiguity annotation errors observed between the two annotators;  \textit{H-Conceptual Misidentification}: conceptual misidentification annotation errors observed between the two annotators; \textit{M-Conceptual(Task), M-Conceptual(Model), M-Boundary(Task), M-Boundary(Model)}: LLM (zero-shot) annotation errors decomposed by error type (conceptual vs. boundary) and source (task-inherent vs. model-specific).}
  \label{fig:perform}
\end{figure*}

\paragraph{\textbf{Effectiveness of Prompting Strategies and Task Inherent Errors}}

Figure~\ref{fig:perform} illustrates the effectiveness of prompting strategies, measured as performance gains over the baseline zero-shot condition, using the best-performing prompt configuration for each task. A negative correlation exists between human annotation accuracy and prompting gains: tasks with a higher level of human alignment with the ground truth tend to exhibit smaller technical improvements. While this trend may initially appear counterintuitive, our error decomposition framework provides an explanation. When we divide the misalignment part of human annotations into boundary ambiguity-oriented errors and conceptual misidentification-oriented errors, we can see that, for example, Uptake, even though it has the lowest human alignment with ground truth, shows a high proportion of boundary ambiguity-oriented errors. These errors theoretically are easier to calibrate by adopting an appropriate technical strategy, such as additional examples that clarify boundary definition. Consistent with this expectation, our results show that incorporating 200 additional human-labeled examples for prompt optimization substantially improves LLM annotation accuracy and yields the best performance on this task. In contrast, MathDial, which has the highest human accuracy, shows very limited improvement from prompting strategies. Our error decomposition indicates that the predominant task-inherent error type for humans is conceptual misidentification, which reflects ambiguity in the construct definition itself. Such ambiguities are inherently more difficult to address through prompting adjustments or the addition of anchor examples, which explains the limited technical improvements observed.

\section{Discussion and Conclusion}
Much of current practice in LLM-based annotation treats reliability as agreement with human-labeled ground truth, implicitly framing those annotations as a technical alignment problem. This alignment-focused perspective overlooks the inherent complexity of annotation tasks that require subjective interpretation and involve unavoidable task ambiguity. In this study, we take a step back to consider the fundamental differences between objective and subjective annotation, and re-examine LLM annotation through the lens of the content analysis approach in the qualitative research paradigm. Extending beyond conventional alignment-based evaluation, we propose a diagnostic LLM annotation evaluation paradigm that incorporates error decomposition to distinguish task-inherent ambiguity from model-driven technical limitations. The paradigm is validated on four public educational annotation tasks to establish its conceptual validity, followed by an exploratory analysis that tests its utility.

Our study offers both theoretical contributions and practical implications. Theoretically, it helps explain a recurring pattern in prior research: LLM-based annotation shows limited and inconsistent performance across social science tasks. Rather than attributing these results solely to model limitations, our results show that they also reflect the inherent ambiguity and limited measurability of many annotation constructs. We provide empirical evidence that such ambiguity is present in human annotations and aligns with human perceptions of task clarity, helping explain why tasks involving higher-order or psychological constructs, such as reasoning, cognitive engagement, pedagogical intention, or metacognitive processes, as well as those requiring context-independent interpretation, remain especially difficult for LLMs to annotate accurately. 

Our study also reveals important limitations of current evaluation practices and shows that the alignment-based evaluation paradigm could be inadequate for subjective annotation tasks.
\textit{From the perspective of error sources},  strict agreement-based metrics could conflate task-inherent ambiguity with model failure, leading to distorted performance assessments that penalize models for differences driven by ambiguous constructs rather than true annotation errors. Moreover, we are concerned that over-pursuing exact matching may create an “overfitting-like” effect, encouraging models to mimic random noise or human bias in the training set rather than capturing the underlying construct. \textit{From the perspective of error types}, the critical question in subjective annotation is whether it captures the underlying construct in ways that remain valid for downstream use. Recent research ~\cite{baumann2025large} shows that even small changes in implementation choices for LLM-based social science annotation can substantially alter research conclusions. Given the increasing use of LLMs for annotation in large-scale social analysis~\cite{handa2025anthropic}, greater caution is needed when such subjective annotations and their downstream analysis carry high-stakes impacts. Our results show that similar overall agreement scores can hide very different types of errors, which differ substantially in how well they capture the intended construct.

Practically, our paradigm provides a lightweight tool for researchers who want to use LLM-based annotation without heavy technical investment or extensive human-labeled data. Across four annotation tasks, we find that estimates of task-inherent error remain stable across different models and prompting strategies. This stability allows researchers to use our diagnostic method in zero-shot settings with a smaller model to approximate an LLM’s performance ceiling and to decide whether further technical optimization is likely to yield meaningful improvements. Beyond performance estimation, error decomposition of model-driven errors could also provide actionable insights to improve annotation quality. For example, when errors are dominated by boundary ambiguity,  adding anchor examples to clarify category boundaries could be the next step, and when conceptual misidentification is prevalent, refining construct definitions in the codebook may be more effective. Taken together, error decomposition reframes conventional evaluation as a diagnostic process that helps identify where further optimization is most likely to be effective.

There are still some limitations within our current work that can inform future research. First, in this initial attempt, we test our paradigm only on ordinal, single-label subjective classification tasks. We plan to further extend this framework to multi-label, non-ordinal tasks as a follow-up in future work and test it in a wider range of social science contexts beyond education.  Second, our approach still relies on the assumption that a reliable ground truth can be established through rigorous human annotation. However, recent research has questioned whether true “gold” labels exist at all, particularly for complex and psychological constructs. We view this concern as closely aligned with our central argument: annotation errors may be inherent to the task itself. Some studies have responded to this challenge by proposing alternative guidelines ~\cite{thomas2025beyond} and process-oriented optimizations ~\cite{guerdan2025validatingllmasajudgesystemsrating}.  Our method cannot resolve this fundamental issue, but we try to adopt a diagnostic perspective to push this discussion forward. Instead of attempting to perfect the ground truth, we focus on estimating the potential risks of deviations from researchers’ annotation intentions and ultimately leave it to researchers to decide whether and how to proceed with LLM-based automation. Lastly, from a technical perspective, a promising future direction involves integrating our framework into the synthetic data generation process for annotation, which could provide richer signals and more informative rewards for reinforcement learning. 

\section*{Appendix}
The appendix can be accessed at: \url{https://github.com/AEQUITAS-Lab/Data-Annotation-LAK-2026}


\bibliographystyle{ACM-Reference-Format}
\bibliography{base}

\end{document}